%% file: neurips_2023.tex
\documentclass{article}

\usepackage[square,numbers]{natbib}

\usepackage[preprint]{neurips_2023}

\usepackage[utf8]{inputenc} 
\usepackage[T1]{fontenc}    
\usepackage{hyperref}       
\usepackage{url}            
\usepackage{booktabs}       
\usepackage{amsfonts}       
\usepackage{nicefrac}       
\usepackage{microtype}      
\usepackage{xcolor}         

\usepackage{hyperref}
\hypersetup{
	colorlinks=true,
	breaklinks=true,
	urlcolor=blue,
	linkcolor=blue,
	bookmarksopen=false,
	filecolor=black,
	citecolor=blue,
	linkbordercolor=blue
}

\input{cmd}

\title{A Note on Knowledge Distillation Loss Function for \\ Object Classification}

\author{%
  Defang Chen\\
  Zhejiang University\\
  \texttt{defchern@zju.edu.cn} \\
}

\begin{document}

\maketitle

\begin{abstract}
	This research note provides a quick introduction to the knowledge distillation loss function used in object classification. In particular, we discuss its connection to a previously proposed logits matching loss function. We further treat knowledge distillation as a specific form of output regularization and demonstrate its connection to label smoothing and entropy-based regularization.
\end{abstract}

\section{Introduction}

Knowledge distillation (KD) was proposed as a model compression technique to distill the knowledge in a powerful yet cumbersome teacher model into a lightweight student model, hoping that it will enhance the student generalization ability \cite{hinton2015distilling}. Since the advent of knowledge distillation, variant approaches have been developed to improve its effectiveness and widen its applications, including feature distillation~\cite{chen2021cross,chen2022knowledge}, collaborative distillation~\cite{chen2020online}, and diffusion-based distillation~\cite{chen2023geometric,zhou2024fast,chen2024trajectory}. In the vanilla KD \cite{hinton2015distilling}, class predictions between teacher and student models are aligned with a newly introduced hyper-parameter\textemdash \textit{temperature} in the softmax activation to control the softness of predicted distributions. In this way, Hinton and his collaborators \cite{hinton2015distilling} argued that the pioneering model compression technique termed logits matching \cite{bucilua2006model} is actually a special case of their proposed approach, provided that the temperature is much higher than the logits in order of magnitude and the logits are zero-mean normalized explicitly. In this note, we provide detailed derivations to review and deepen our understanding of this connection. 
We first prove that with a single \textit{infinity temperature} condition, we could already build the connection between these two loss functions, although attached with an extra regularization. We further point out that \textit{equal-mean normalization} for logits is enough to establish the exact equivalence. Finally, we discuss knowledge distillation from the output regularization perspective.

\section{Preliminary}

We first briefly recap the basic concept of object classification using deep neural networks, especially from the perspective of knowledge distillation. Then, a formal description of the standard knowledge distillation and logits matching loss functions are introduced with necessary notations.

\subsection{Distilling Knowledge from Human Labellings}

Given a training dataset $\calD=\{(\bfx, \bfy)\}$ consisting of $n$ objects from $k$ categories, we denote the category label of the object $\bfx$ as $\bfy$. 
Take one-class image classification problem as an example, we have $\bfx \in \bbR^{c\times h \times w}$, where $c$ denotes the channel dimension, $h$ and $w$ denote the spatial dimensions. Besides, only one element in the vector $\bfy\in \bbR^k$ equals one and the other elements all equal zero (e.g., $\bfy_i=1$, and $\bfy_j=0$, $\forall j\neq i$ if this object belongs to the $i$-th category). 
We then train a deep neural network with the parameter $\bftheta$ to learn a mapping from the object space to the category space. 

Given the object $\bfx$, we denote the unnormalized prediction of a deep neural network as $\bfz$, which is also known as \textit{logits}, and denote its softmax-based normalized version as $\bfp$, which is also known as \textit{class predictions}. Mathematically, we have $\bfp_{i}=\exp\left(\bfz_{i}\right)/\sum_{j=1}^{k}\exp \left(\bfz_{j}\right)$, where $\bfz_{i}$ and $\bfp_i $ denote the $i$-th element of $\bfz$ and $\bfp$, respectively.
The model parameter $\bftheta$ is randomly initialized and updated using the following cross-entropy loss function:
\begin{equation}
	\calL_{\mathrm{CE}}(\bfy, \bfp) = - \sum_{i=1}^{k} \bfy_i \log \bfp_i, \quad \text{or equivalently}, \quad \calL_{\mathrm{KL}}(\bfy, \bfp) = \sum_{i=1}^{k} \bfy_i \log \frac{\bfy_i}{\bfp_i},
\end{equation}
with the gradient as follows
\begin{equation}
	\frac{\partial \calL_{\mathrm{CE}}}{\partial \bfp_{j}}=-\frac{\bfy_{j}}{\bfp_{j}},\qquad \frac{\partial \bfp_{j}}{\partial \bfz_{i}} = 
	\begin{cases}
		\bfp_{j}(1-\bfp_{j})		
		& \mbox{$i=j$}		\\
		-\bfp_{i}\bfp_{j}		  
		& \mbox{$i\neq j$}.
	\end{cases} 
\end{equation}
Then, we take the partial derivative of $\calL_{\mathrm{CE}}$ with respect to $\bfz_{i}$
\begin{equation}
	\begin{aligned}	
		\frac{\partial \calL_{\mathrm{CE}}}{\partial \bfz_{i}}
		&=
		\sum_{j}\frac{\partial \calL_{\mathrm{CE}}}
		{\partial \bfp_{j}}\frac{\partial \bfp_{j}}
		{\partial \bfz_{i}}=-\sum_{j}\frac{\bfy_{j}}{\bfp_{j}}\frac{\partial \bfp_{j}}
		{\partial \bfz_{i}}
		=-\left[\underbrace{\frac{\bfy_{j}}{\bfp_{j}}\bfp_{j}\left(1-\bfp_{j}\right)}_{j=i}+\underbrace{\sum_{j}\frac{\bfy_{j}}{\bfp_{j}}\left(-\bfp_{i}\bfp_{j}\right)}_{j\neq i}\right]\\
		&=-\left[\underbrace{\bfy_{j}\left(1-\bfp_{j}\right)}_{j=i}+\underbrace{\sum_{j}\bfy_{j}\left(-\bfp_{i}\right)}_{j\neq i}\right]
		=-\left[\bfy_{i}-\bfp_{i}\sum_{j}\bfy_{j}\right]
		=\bfp_{i}-\bfy_{i}.
	\end{aligned}
\end{equation}
\begin{remark}
	The standard learning objective for object classification can be interpreted as distilling knowledge from human labellings, where the regular model acts as a ``student'' model and the ground-truth label for each object acts as the output of a(n) ``teacher/oracle'' model.  
\end{remark}

\textbf{Label Smoothing}:
To alleviate the over-fitting issue in model training or the mis-labeling issue in data collection, label smoothing \cite{szegedy2016rethinking} technique was proposed. It replaces the hard label $\bfy$ as the smoothed label $\bfy^{\prime}$, with $\bfy^{\prime}=(1-\alpha)\bfy + \alpha \bfu$ and $\bfu_i=1/k$, $\forall i$. In this case, the above loss function becomes\footnote{We omit the constant term irrelevant to the parameter optimization.}
\begin{equation}
	\label{eq:ls}
	\calL_{\mathrm{LS}}(\bfy^{\prime}, \bfp) = (1-\alpha) 	\calL_{\mathrm{CE}}(\bfy, \bfp) + \alpha \calL_{\mathrm{KL}}(\bfu, \bfp).
\end{equation}
\textbf{Confidence Penalty}: Another similar output-based regularization reverses the KL term in the above equation, aiming to penalize a low-entropy model prediction~\cite{pereyra2017regularizing} 
\begin{equation}
	\label{eq:cp}
	\calL_{\mathrm{CP}}(\bfy^{\prime}, \bfp) = (1-\alpha) 	\calL_{\mathrm{CE}}(\bfy, \bfp) + \alpha \calL_{\mathrm{KL}}(\bfp, \bfu).
\end{equation}
The above two loss functions can be unified using skew-Jensen divergence~\cite{nielsen2011burbea,meister2020generalized}. 

\subsection{Distilling Knowledge from Neural Networks}

We next denote class predictions of the teacher model with the parameter $\bftheta^t$ and the student model with the parameter $\bftheta^s$ as $\bfp^{t}$ and $\bfp^{s}$, which are produced by normalized logits $\bfz^{t}$ and $\bfz^{s}$, respectively.

A straightforward approach to distill knowledge from a pre-trained powerful teacher model to a simple student model is adopting the following logits matching loss function~\cite{bucilua2006model,ba2014deep} to update the student model's parameter $\bftheta^s$: 
\begin{equation}
	\label{eq:logit}
	\begin{aligned}
		\calL_{\mathrm{LM}}(\bfz^{t}, \bfz^{s})=\frac{1}{2k}\|\bfz^{t} - \bfz^{s}\|^{2}_{2}=\frac{1}{2k}\sum_{i=1}^{k}\left(\bfz^{t}_{i}-\bfz^{s}_{i}\right)^2,
	\end{aligned}
\end{equation}
with the gradient $\partial \calL_{\mathrm{LM}}/\partial \bfz^{s}_{i}=\left(\bfz^{s}_{i}-\bfz^{t}_{i}\right)/k$, and the gradient $\partial  \calL_{\mathrm{LM}}/\partial \bftheta^s=\partial  \calL_{\mathrm{LM}}/\partial \bfz^s \cdot \partial \bfz^s / \partial \bftheta^s$.

Similarly, we can also adopt the vanilla knowledge distillation loss function~\cite{hinton2015distilling} for knowledge transfer. This approach introduces temperature $\tau$ as a hyper-parameter to soften the model-predicted probability distributions, and modifies the relationship between logits $\bfz$ and predictions $\bfp$ as follows
\begin{equation}
	\label{eq:softmax}
	\bfp^{t}_{i}=\frac{\exp\left(\bfz^{t}_{i}/\tau\right)}{\sum_{j=1}^{k}\exp \left(\bfz^{t}_{j}/\tau\right)}, \qquad \bfp^{s}_{i}=\frac{\exp\left(\bfz^{s}_{i}/\tau\right)}{\sum_{j=1}^{k}\exp \left(\bfz^{s}_{j}/\tau\right)}.
\end{equation}
The vanilla knowledge distillation loss function is 
\begin{equation}
	\label{eq:kd}
	\begin{aligned}
		\calL_{\mathrm{KD}}(\bfp^{t}, \bfp^{s})=\sum_{i=1}^{k}\bfp^{t}_{i}\log\frac{\bfp^{t}_{i}}{\bfp^{s}_{i}}=-\sum_{i=1}^{k}\bfp^{t}_{i}\log \bfp^{s}_{i}+\sum_{i=1}^{k}\bfp^{t}_{i}\log \bfp^{t}_{i}.
	\end{aligned}
\end{equation}
The second term in Equation~(\ref{eq:kd}) is a negative entropy of $\bfp^{t}$, which is  irrelevant to the update of student's parameters.
Similarly, we take the partial derivative of $\calL_{\mathrm{KD}}$ w.r.t. $\bfz^{s}_{i}$:
\begin{equation}
	\frac{\partial \calL_{\mathrm{KD}}}{\partial \bfp^{s}_{j}}=-\frac{\bfp^t_{j}}{\bfp^s_{j}},\qquad \frac{\partial \bfp^s_{j}}{\partial \bfz^s_{i}} = 
	\begin{cases}
		\frac{1}{\tau}\bfp^s_{j}\left(1-\bfp^s_{j}\right)		
		& \mbox{$i=j$}		\\
		-\frac{1}{\tau}\bfp^s_{i}\bfp^s_{j}		  
		& \mbox{$i\neq j$}.
	\end{cases} 
\end{equation}
\begin{equation}
	\label{eq:gradient_kd}
	\begin{aligned}	
		\frac{\partial \mathcal{L}_{\mathrm{KD}}}{\partial \bfz^s_{i}}
		=
		\sum_{j}\frac{\partial \mathcal{L}_{\mathrm{KD}}}
		{\partial \bfp^s_{j}}\frac{\partial \bfp^s_{j}}
		{\partial \bfz^s_{i}}=-\sum_{j}\frac{\bfp^t_{j}}{\bfp^s_{j}}\frac{\partial \bfp^s_{j}}
		{\partial \bfz^s_{i}}
		=\frac{1}{\tau}\left(\bfp^s_{i}-\bfp^t_{i}\right).
	\end{aligned}
\end{equation}

\begin{remark}
	The standard knowledge distillation loss function (Equation~(\ref{eq:kd})) multiplying $\tau^2$ acts as a regularized logits matching loss function (Equation~(\ref{eq:logit})) under the infinity temperature.
\end{remark}
\begin{proof}
We then prove the equivalence between $\tau^2\calL_{\mathrm{KD}}$ and $\calL_{\mathrm{LM}}$ \cite{hinton2015distilling,kim2021comparing} below. 
\begin{equation}
	\label{eq:kd_grad}
	\begin{aligned}
		\lim_{\tau \rightarrow \infty}\tau^2\frac{\partial \calL_{ \mathrm{KD}}}{\partial \bfz^{s}_{i}}
		&\overset{\textcircled{\small{1}}}{=}\lim_{\tau \rightarrow \infty}\tau\left(\bfp^{s}_{i}-\bfp^{t}_{i}\right)=		\lim_{\tau \rightarrow \infty}\tau\left(\frac{\exp\left(\bfz^{s}_{i}/\tau\right)}{\sum_{j=1}^{k}\exp \left(\bfz^{s}_{j}/\tau\right)}-\frac{\exp\left(\bfz^{t}_{i}/\tau\right)}{\sum_{j=1}^{k}\exp \left(\bfz^{t}_{j}/\tau\right)}\right)\\
		&\overset{\textcircled{\small{2}}}{=}\lim_{\tau \rightarrow \infty} \left(
		\frac{\sum_{j=1}^{k} \tau \left(\exp\left(\left(\bfz^t_j - \bfz^t_k\right)/\tau\right)-1\right)-\sum_{j=1}^{k} \tau \left(\exp\left(\left(\bfz^s_j - \bfz^s_k\right)/\tau\right)-1\right)}
		{\left(\sum_{j=1}^{k} \exp\left(\left(\bfz^t_j - \bfz^t_k\right)/\tau\right)\right)\left(\sum_{j=1}^{k} \exp\left(\left(\bfz^s_j - \bfz^s_k\right)/\tau\right)\right)}
		\right)
		\\
		&=\frac{1}{k}\left(\bfz^{s}_{i}-\bfz^{t}_{i}\right)-\frac{1}{k^2}\sum_{j=1}^{K}\left(\bfz^{s}_{j}-\bfz^{t}_{j}\right)\\
	\end{aligned}
\end{equation}
For $\textcircled{\small{1}}$, the detailed derivation is provided in Equation~(\ref{eq:gradient_kd}), and we substitute the Equation~(\ref{eq:softmax}) into $\bfp^{s}_{i}$ and $\bfp^{t}_{i}$; 
for $\textcircled{\small{2}}$, the detailed derivation is provided in Section A.2 of \cite{kim2021comparing}, and we require $\tau$ goes to positive infinity to leverage the Taylor approximation of exponential function. 
Therefore, the standard knowledge distillation loss function acts as a regularized logits matching loss function:
\begin{equation}
	\calL_{\mathrm{LM_r}} = \frac{1}{2k}\|\bfz^{t} - \bfz^{s}\|^{2}_{2} - \frac{1}{2k^2} \left(\sum_{j=1}^{k} \bfz^s_j - \sum_{j=1}^{k}\bfz^t_j \right)^2 +\mathrm{Constant}.
\end{equation}
The above loss function also implies that when the summation of logits predicted by teacher and student models are equal, i.e, $\sum_{j}^{k}\bfz^{s}_{j}=\sum_{j}^{k}\bfz^{t}_{j}$, the extra regularization term will just disappear. 
\end{proof}
\begin{remark}	
	Based on the infinity large temperature and equal-mean normalization for logits (i.e., $\sum_{j}^{k}\bfz^{s}_{j}=\sum_{j}^{k}\bfz^{t}_{j}$), the gradient of $\tau^2\calL_{\mathrm{KD}}$ equals to that of $\calL_{\mathrm{LM}}$ and thus we can conclude that the effects of these two losses are exactly the same. 
\end{remark}
%

\subsection{Knowledge Distillation as Output Regularization}

In practice, the knowledge distillation loss function is generally used together with the original cross-entropy loss function. That is to say, we adopt the following loss function to optimize parameters of a student model:
\begin{equation}
	\label{eq:kd_ce}
	\calL_{ \mathrm{KD}^{\prime}} = (1-\alpha) 	\calL_{ \mathrm{CE}}(\bfy, \bfp^{s}) + \alpha \calL_{ \mathrm{KD}}(\bfp^{t}, \bfp^{s}). 
\end{equation}
Comparing the loss function in Equation~(\ref{eq:ls}), we conclude that class predictions of the teacher model act as adaptive label smoothing to prevent the student output being over-confident~\cite{yuan2020revisiting}.

\begin{remark}
	The knowledge distillation loss function in Equation~(\ref{eq:kd_ce}) acts as an adaptive label smoothing loss function in Equation~(\ref{eq:ls}).
\end{remark}

\bibliographystyle{alpha}
\bibliography{mybibfile}
\end{document}

%% file: cmd.tex
\usepackage{amsmath}
\usepackage{amssymb}
\usepackage{amsthm}
\usepackage{booktabs}
\usepackage{xspace}
\usepackage{graphicx}

\usepackage{float}
\usepackage{subcaption}

\usepackage{mathtools}
\usepackage{pifont}
\usepackage{cancel}
\usepackage{algorithm}
\usepackage{algorithmic}
\usepackage{multirow}
\usepackage{colortbl}
\usepackage{utfsym}

 \newtheorem{remark}{Remark}

\def\bfp{\mathbf{p}}

\def\bfu{\mathbf{u}}

\def\bfx{\mathbf{x}}
\def\bfy{\mathbf{y}}
\def\bfz{\mathbf{z}}

\def\bbR{\mathbb{R}}

\def\calL{\mathcal{L}}
\def\calD{\mathcal{D}}

\def\bftheta{\boldsymbol{\theta}}

\def\eqref#1{(\ref{#1})}
